# Pancreas Segmentation with Probabilistic Map Guided Bi-directional Recurrent UNet

**Jun Li[1], Xiaozhu Lin[2], Hui Che[3], Hao Li[1] and Xiaohua Qian[1]**

[1] School of Biomedical Engineering, Shanghai Jiao Tong University, Shanghai 200030, China.
[2] Department of Radiology, Ruijin Hospital, Shanghai Jiao Tong University School of Medicine, Shanghai, China.
[3] Biomedical Engineering Department, Rutgers University, New Jersey 08901, USA.

Correspondence e-mail: xiaohua.qian@sjtu.edu.cn

## Abstract

Pancreas segmentation in medical imaging is of great significance for clinical pancreas diagnostics and treatment. However, the large population variations in the pancreas shape and volume cause enormous segmentation difficulties, even for state-of-the-art algorithms utilizing fully convolutional neural networks (FCNs). Specifically, pancreas segmentation suffers from the loss of statement temporal information in 2D methods, and the high computational cost of 3D methods. To alleviate these problems, we propose a probabilistic-map-guided bi-directional recurrent UNet (PBR-UNet) architecture, which fuses intra-slice information and inter-slice probabilistic maps into a local 3D hybrid regularization scheme, which is followed by a bi-directional recurrent optimization scheme. The PBR-UNet method consists of an initial estimation module for efficiently extracting pixel-level probabilistic maps and a primary segmentation module for propagating hybrid information through a 2.5D UNet architecture. Specifically, local 3D information is inferred by combining an input image with the probabilistic maps of the adjacent slices into multi-channel hybrid data, and then hierarchically aggregating the hybrid information of the entire segmentation network. Besides, a bi-directional recurrent optimization mechanism is developed to update the hybrid information in both the forward and the backward directions. This allows the proposed network to make full and optimal use of the local context information. Quantitative and qualitative evaluation was performed on the NIH Pancreas-CT and MSD pancreas dataset, and our proposed PBR-UNet method achieved similar segmentation results with less computational cost compared to other state-of-the-art methods.

Keywords: pancreas segmentation, bi-directional recurrent UNet, medical image segmentation

## 1. Introduction

Accurate segmentation of the human pancreas in medical imaging data is an essential prerequisite for relevant medical image analysis and surgical navigation systems. However, pancreas segmentation is quite challenging due to the considerable variations in the shape, and the pancreas' vulnerability to elastic deformations resulting from breathing and heartbeat. Therefore, developing a robust and accurate pancreas segmentation model is of profound significance for performance improvement in computer-assist surgery techniques.

Nowadays, developing satisfactory methods for pancreas segmentation is still challenging. Compared with some other human organs such as the heart and liver, the pancreas exhibits a higher anatomical variability (Zhou et al., 2017), as shown in Fig. 1. Variations in the background tissues and dramatic volume changes can undermine the performance of any start-of-the-art method for pancreas segmentation (Roth et al., 2015a). Consequently, the pancreas has been typically

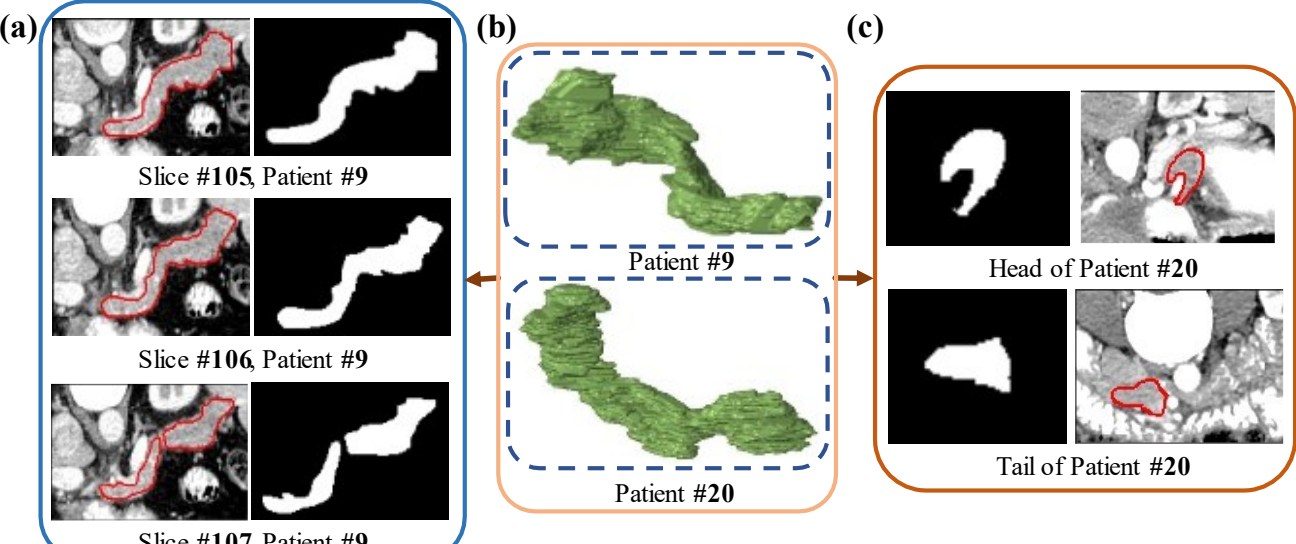

Fig. 1. Examples of pancreas scan: (a) Three adjacent slices with high correlation; (b) 3D pancreas data from two subjects showing different spatial shapes; (c) A small correlation exists between the shapes of the pancreas head and tail.

considered among the most complex organs for segmentation (Roth et al., 2015b).

Pancreas segmentation methods can be roughly divided into two categories, namely methods based on top-down multi-atlas registration and label fusion (MALF) (Tong et al., 2015, Karasawa et al., 2017, Oda et al., 2016), and methods based on deep learning (Roth et al., 2015a, Roth et al., 2015b, Zhou et al., 2017). For a MALF method, volumetric multiple-atlas registration is combined with a robust label-fusion scheme to optimize the per-pixel pancreas segmentation (Fu et al., 2018). Due to the high shape variability and blurred boundaries of the pancreas, the accuracies of MALF-based methods on benchmark datasets range merely from 69.6% to 78.5%. Nevertheless, deep learning methods, such as FCN (Long et al., 2015), UNet (Ronneberger et al., 2015), and DeepLab (Chen et al., 2018) have been demonstrated to be powerful tools for organ segmentation thanks to the availability of large annotated datasets and computational resources compared with traditional segmentation techniques.

Many existing deep-learning pancreas segmentation methods are based on 2D data (Fu et al., 2018, Cai et al., 2017, Zhou et al., 2017). Specifically, 2D segmentation methods process each slice of a volume as a separate input. Then, the segmentation results of all slices are combined to construct a 3D pancreas object. Because three non-identical views give quite different images of the pancreas, (Zhou et al., 2017) proposed training three 2D models for segmentation in each of the coronal, sagittal, and axial views, and then merged these three results via majority voting to produce final segmentation. Although this method can obtain rough spatial shapes, it ignores the spatial continuity of the pancreas. (Fu et al., 2018) proposed a richer convolutional feature network, which performs pancreas segmentation by extracting multiscale information via a multi-layer up-sampling structure. Although this method enhances the extraction of intra-slice information, it completely ignores the 3D information. In general, 2D networks process each slice independently, causing it to ignore the relationship between adjacent slices (Li et al., 2018). This impedes the extraction of contextual features and restricts the pancreas segmentation performance.

Since 2D networks cannot capture contextual information, 3D networks were proposed by using the CT volume as the network input (Fang et al., 2019, Zhu et al., 2018, Oktay et al., 2018, Guo et al., 2018, Liu et al., 2018). However, these 3D networks require significant computational and memory resources and hence rely excessively on high-performance servers (Li et al., 2018). (Oktay et al., 2018) proposed an attention gate model, based on a 3D UNet architecture. While this model can suppress irrelevant regions and highlight useful salient features, the model suffers from large computational costs. To relieve the pressure of a global fine-grained search for the pancreas, (Roth et al., 2018b) proposed using a two-stage segmentation scheme, in which the second 3D FCN has reduced computations and is focused on the segmentation of the target organ. However, the sub-volume approach they adopted may cause segmentation discontinuities or inconsistencies at overlapping window boundaries (Roth et al., 2018a). The performance and costs of a complete 3D convolutional architecture were further investigated (Lai, 2015). It was found that such 3D architecture provided slightly better performance in comparison to 2D methods but caused a significant and disproportionate increase in computing costs (2D: 51.56 mins, 3D: 173.73 mins) (Lai, 2015). The requirement of a high-memory footprint limits two performance improvement factors, namely the network depth and the filter field of view (Simonyan and Zisserman, 2014). So, the basic 3D networks can hardly achieve satisfactory performance in pancreas segmentation.

Two-stage learning frameworks for coarse-to-fine pancreas segmentation were also proposed to overcome the susceptibility of one-pass learning strategies to background interference (Zhu et al., 2018, Roth et al., 2018b, Roth et al., 2018a, Zhou et al., 2017). In such frameworks, the region of interest (ROI) is roughly localized by the initial segmentation, and then a finer segmentation is carried out by focusing on the localized ROI. However, (Yu et al., 2018) presented two failure cases using the coarse-to-fine scheme (Zhou et al., 2017) and indicated that the fine-scale segmentation deteriorated in some cases due to the lack of contextual information.

The success of the long short-term memory (LSTM) networks is essentially due to their effective consideration of long-span dependencies, and contextual information. In some studies (Cai et al., 2017, Hua et al., 2019, Yang et al., 2018), LSTM networks were applied to directly embed contextual information as time series into models. But in the LSTM-based segmentation framework, the LSTM module usually serves as a single refinement module following the main segmentation networks (Yang et al., 2018). Indeed, the LSTM makes limited segmentation improvements, enlarges the entire network, and requires more computing resources.

The pancreas globally shows a significant anatomical variability, while it locally exhibits strong morphology and pattern correlation among adjacent CT slices (Fig. 1a). These global and local characteristics demonstrate that the local 3D information or inter-slice information is critical for developing a segmentation model with high precision and robustness (Gonzalez et al., 2019). However, while earlier segmentation models exploited 2D features or global 3D information, few models accounted for the local 3D context. In general, multi-channel networks have achieved better results than single-channel networks without a significant increase in the computational burden (Lai, 2015).

Motivated by the above observations, we developed a novel pancreas segmentation model based on local 3D hybrid information and a bi-directional recurrent 2.5D UNet architecture, namely the probabilistic-map-guided bi-directional recurrent UNet (PBR-UNet). In this model, the original map of a CT slice is combined with probabilistic maps of the adjacent slices to infer local 3D hybrid information that can be used for guiding the segmentation of the centre slice. This information is propagated in the 2.5D UNet and then optimized through a bi-directional recurrent scheme to improve and refine the segmentation results. Specifically, we firstly apply an initial estimation model to extract, for each slice, a pixel-level probabilistic map, which represents the per-pixel probability of belonging to the pancreas. Then, the initial probabilistic maps of the adjacent slices are combined with the map of the centre slice into hybrid multi-channel data, which contains local 3D hybrid information of the centre slice. Under the constraints of the local 3D context, the segmentation of the centre slice could be constrained, resulting in stable results. Finally, a bi-directional recurrent scheme is applied to the primary segme Following previous studies on MSD, we combine the pancreas and tumor to belong to a single target foreground in our experiments。 ntation to optimize the local 3D hybrid information. We use each primary segmentation output to update the probabilistic maps in the hybrid multi-channel data, make the local 3D hybrid information more precise, and boost the final segmentation performance.

In summary, our PBR-UNet framework has the following two technical contributions:

1) Introducing local 3D hybrid information. A probabilistic-map-guided segmentation model is developed to combines intra-slice information and probabilistic maps of adjacent slices to form the local 3D hybrid information. The proposed model balances the requirements for high efficiency in statement temporal information utilization and low computational costs, thus avoids the problems of lack of context in 2D models, and the impact of anatomical variability or computational costs on the 3D ones.
2) Constructing a bi-directional recurrent scheme. A bi-directional recurrent update scheme is proposed to optimize local 3D hybrid information in 2.5D UNet. This information is propagated and updated in both the forward and backward directions to make full use of the local context. Under the guidance of the optimized local contextual information, the burden of searching for the optimal pancreas segmentation is well-relieved, and high-precision results can be achieved. Besides, this bi-directional recurrent update scheme can be embedded in most segmentation models.

The remainder of this paper is organized as follows. Section 2 gives the details of the methods used in our proposed model. Section 3 presents the experimental results, Section 4 discusses our findings, and Section 5 highlights key conclusions.

## 2. Materials and methods

Figure 2 shows the flowchart of the PBR-UNet framework for pancreas segmentation. We begin with an introduction of the data used in this work (Sec. 2.1) and a problem definition (Sec. 2.2), followed by a detailed description of the initial estimation (Sec. 2.3) and primary segmentation (Sec. 2.4) stages. Then, we summarize our basic architecture and parameter setting (Sec. 2.5). Finally, we present the inference schemes (Sec. 2.6) and evaluation metrics (Sec. 2.7).

### *2.1. Dataset and pre-processing*

Following most of the earlier approaches, we use the NIH Pancreas-CT dataset (Roth et al., 2015b) to extensively and quantitatively evaluate our proposed model. This dataset contains the abdominal CT scans of 82 patients where each scan has a size of $512 \times 512 \times L$, and $L \in [181,466]$ is the number of slices in each CT scan. We empirically truncated the CT radio-density values to the range of [-100, 240] HU and normalized them to have a zero mean and unit variance. A 4-fold cross-validation (CV) scheme was used, we split the dataset into 4 fixed folds. Every time, we train the models on 3 out of 4 subsets and test them independently on the remaining one and all parameters remain the same. Finally, we report the average of the four folds. Besides, we also evaluate our method on the pancreas Medical Segmentation Decathlon (MSD) challenge dataset (Simpson et al., 2019), which contain 281 CT scans with labelled pancreas and pancreas tumor. Each scan has a size of $512 \times 512 \times L_{msd}$, and $L_{msd} \in [37,751]$ is the number of slices in each CT scan. Following previous studies on MSD, we combine the pancreas and tumor into a single target foreground in our experiments. The data pre-processing steps of MSD are consistent with NIH.

### *2.2. Problem definition*

In this section, we formulate the problem of pancreas segmentation from 3D CT scans in terms of basic mathematical notations. Let $I \in R^{m \times h \times w}$ be the 3D scan data of a patient, where $m$ denotes the total number of slices, $h$ and $w$ refer to the slice height and width, respectively. The

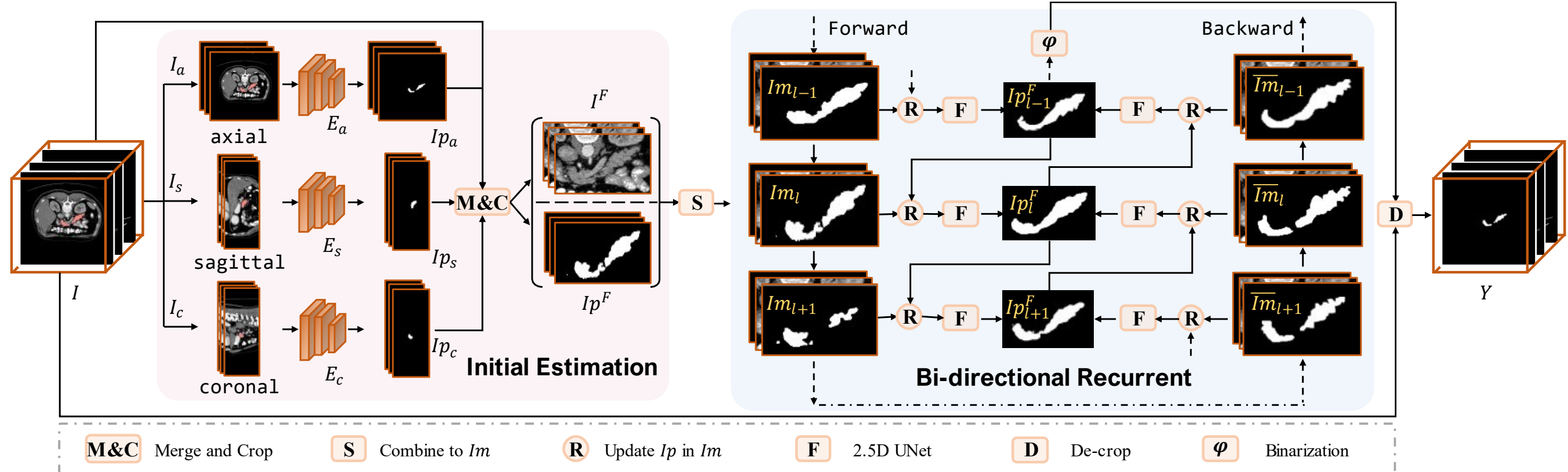

Fig. 2. Illustration of the PBR-UNet pipeline for pancreas segmentation.

annotation of $I$ is a binary segmentation mask $\hat{Y} \in R^{m\times h\times w}$, and it is defined as:

$$\hat{Y}_{m,h,w} = \begin{cases} 0, & I(m,h,w) \notin pancreas \\ 1, & I(m,h,w) \in pancreas \end{cases} \quad (1)$$

where a value of 1 means that the voxel belongs to the pancreas, 0 means that the voxel belongs to the background. The mapping function $\Omega$ is constructed based on the ground truth $\hat{Y}$, and this function outputs the pixel-wise segmentation maps $Y$ for given 3D scan $I$. The mapping function should be constructed such that the similarity between $Y$ and the ground truth map $\hat{Y}$ is as high as possible. $\Omega$ can be written as the composition of two functions: $E$ and $F$. The function $E$ returns the initial segmentation estimate, *i.e.*, the pixel-wise probabilistic maps $Ip$ of the 3D scan $I$. Next, $Ip$ and $I$ are cropped with a rectangle, based on the binarized $Ip$, to get $Ip^F$ and $I^F$. Then, $Ip^F$ and $I^F$ are combined as hybrid multi-channel data $Im$ for bi-directional recurrent updates during the primary segmentation. The function $F$ returns the primary segmentation outputs resulting from bi-directionally propagating and optimizing local 3D hybrid information. To spread and make full use of the local context information, we design the primary segmentation $F$ in both forward and backward directions, which can be expressed as $F = backward\ (forward\ (Im))$, and the design details will be given in the next sub-section. Thus, the pancreas segmentation problem can be roughly formulated as the problem of minimizing the loss function:

$$L(E,F) = \min\left( \frac{1}{m} \sum_{i=1}^{m} \left( F\left(E(I_{i-1}), I_i, E(I_{i+1})\right) - \hat{Y}_i \right)^2 \right) \quad (2)$$

### 2.3. Initial estimation of probabilistic maps

In the first stage of the proposed PBR-UNet framework, intra-slice features are extracted, and the probabilistic map is obtained for each slice using the initial estimation module. To utilize the volume information, we use a multi-view FCN (M-FCN) (Long et al., 2015) to quickly obtain pixel-wise probabilistic maps. The M-FCN ($E$) contains three FCN models for conducting segmentation along the coronal ($E_c$), sagittal ($E_s$), and axial ($E_a$) views. These models have the same structure and train separately with data associated with their respective views. Specifically, the 3D data $I \in R^{m\times h\times w}$ is slices along three axes to obtain 2D slices of the corresponding view, *i.e.*, $I_c \in R^{w\times m\times h}$, $I_s \in R^{h\times m\times w}$, and $I_a \in R^{m\times h\times w}$. The slices from different views are feed into the corresponding model to obtain the segmentation probabilistic maps, this process can be expressed as:

$$Ip_{axes} = E_{axes}(I_{axes}), \quad \|Ip_{axes}\| \in [0,1] \quad (3)$$

Then, an arithmetic averaging is applied to merge these three results:

$$Ip = \frac{1}{3}(Ip_c, Ip_s, Ip_a), \quad \|Ip\| \in [0,1] \quad (4)$$

The probabilistic maps obtained by fusion come from the combined judgment of the three views, which makes them have a better confidence level. This property is beneficial to reduce the interference of the probabilistic maps to the recurrent segmentation process. On the other hand, the probabilistic maps $Ip$ obtained from the initial estimation can also provide a coarse region of interest (ROI) of each volume data. Thus, probabilistic maps are employed to provide a shrunk image region for the bi-directional recurrent segmentation. Specifically, we first empirically employ 0.5 as a threshold $\varepsilon$ to binarize the probabilistic maps:

$$Z_{Ip} = \varphi(Ip;\varepsilon) = \begin{cases} 1, & probability\ value \geq \varepsilon \\ 0, & probability\ value < \varepsilon \end{cases} \quad (5)$$

Based on the minimum bounding box obtained from the binarized probabilistic maps, the original image $I$ and probabilistic maps $Ip$ are cropped with a rectangle derived from $Z_{Ip}$:

$$I^F, Ip^F = C(I, Ip; Z_{Ip}, m) \quad (6)$$

The function $C$ represents the cropping of $I$ and $Ip$ based on the bounding box obtained from $Z_{Ip}$. $m$ represents the padding margin used to ensure that the cropping does not damage the integrity of the pancreas region, which is set to 20 in our experiments.

### *2.4. Bi-directional recurrent segmentation*

Due to the volumetric continuity of the pancreas, experienced radiologists typically localize the pancreas in a CT slice according to the adjacent slices. However, the 2D models can only capture intra-slice features, 3D networks suffer from dramatic anatomical variability. Thus, we propose to leverage the probabilistic maps of adjacent slices to guide the segmentation process with local 3D hybrid information. In this sub-section, we present details of combining multi-channel data and primary segmentation with the bi-directional recurrent scheme. Specifically, we first take three consecutive slices as an example to introduce the process of constructing the hybrid multi-channel data $Im$ based on the probabilistic maps $Ip^F$ and original image $I^F$. With this hybrid multi-channel data, we then present the details of the bi-directional recurrent segmentation process.

#### *2.4.1. Build hybrid multi-channel data*

To fuse the intra-slice and context information into local 3D hybrid information, each center slice is combined with the probabilistic maps of its adjacent slices into hybrid multi-channel data $Im$, which can guide the bi-directional recurrent segmentation process. Specifically, hybrid multi-channel data $Im$ is formed by combining the estimated probabilistic map $Ip^F$ and the original data $I^F$ using the transformation $S$:

$$Im = S\left(I^F, Ip^F\right) = \left[\cdots, \left(Ip^F_{l-1}, I^F_l, Ip^F_{l+1}\right), \cdots\right] \tag{7}$$

where $l$ denotes the $l$-th slice in the 3D data. Fig. 3 describes this transformation process. Since the first slice does not have a preceding slice and the last slice does not have a succeeding slice, they are duplicated.

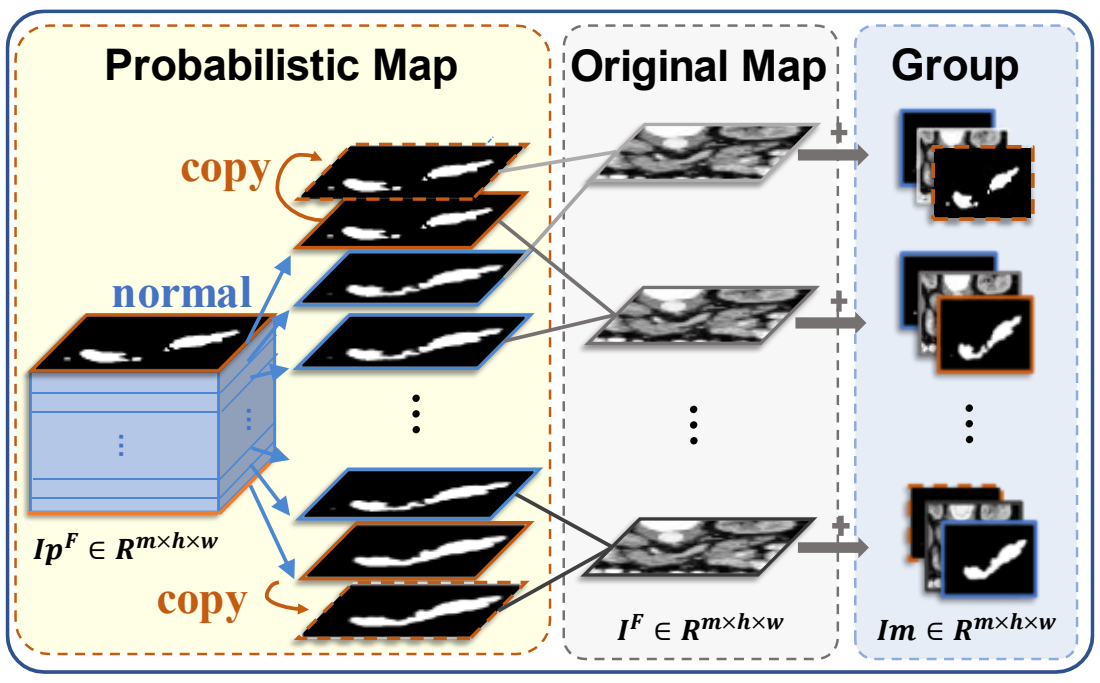

Fig. 3. Construction of the local 3D hybrid information by combining the original map and the probabilistic map to form hybrid multi-channel data (in this case, 3 channels are used). To maintain the data integrity, the first and last slices are duplicated.

#### *2.4.2. Bi-directional recurrent scheme*

The local 3D hybrid information in the multi-channel data can be propagated through the 2.5D UNet to guide the primary segmentation. Moreover, the output of the 2.5D UNet are used to update the corresponding part of the hybrid multi-channel data. The final output is obtained after the bi-directional recurrent update and binarization steps. Fig. 4 illustrates the bi-directional recurrent update scheme with multi-channel data, where three channels are used for illustration. The bi-directional recurrent update scheme consists of two main steps: segmentation and updating. For segmentation, as described above, $I^F_l$, $Ip^F_{l-1}$ and $Ip^F_{l+1}$ form a hybrid three-channel data sample $Im_l$, which is fed into the 2.5D UNet (indicated by $F$) to output a new probabilistic map $\overline{Ip}^F_l$ of the $l$-th slice. For updating, since the hybrid multi-channel data $Im_{l-1}$ and $Im_{l+1}$ both contain $Ip^F_l$, a one-way update would make this strategy benefit from statement temporal information in only one direction. For instance, if only the forward flow in Fig. 4 was preserved, the update of $Im_l$ would only be relevant to its previous data $Im_{l-1}$. However, the update of $Im_l$ should be relevant to both its previous and next data ($Im_{l-1}$ and $Im_{l+1}$), so we design the recurrent update mechanism to be bi-directional.

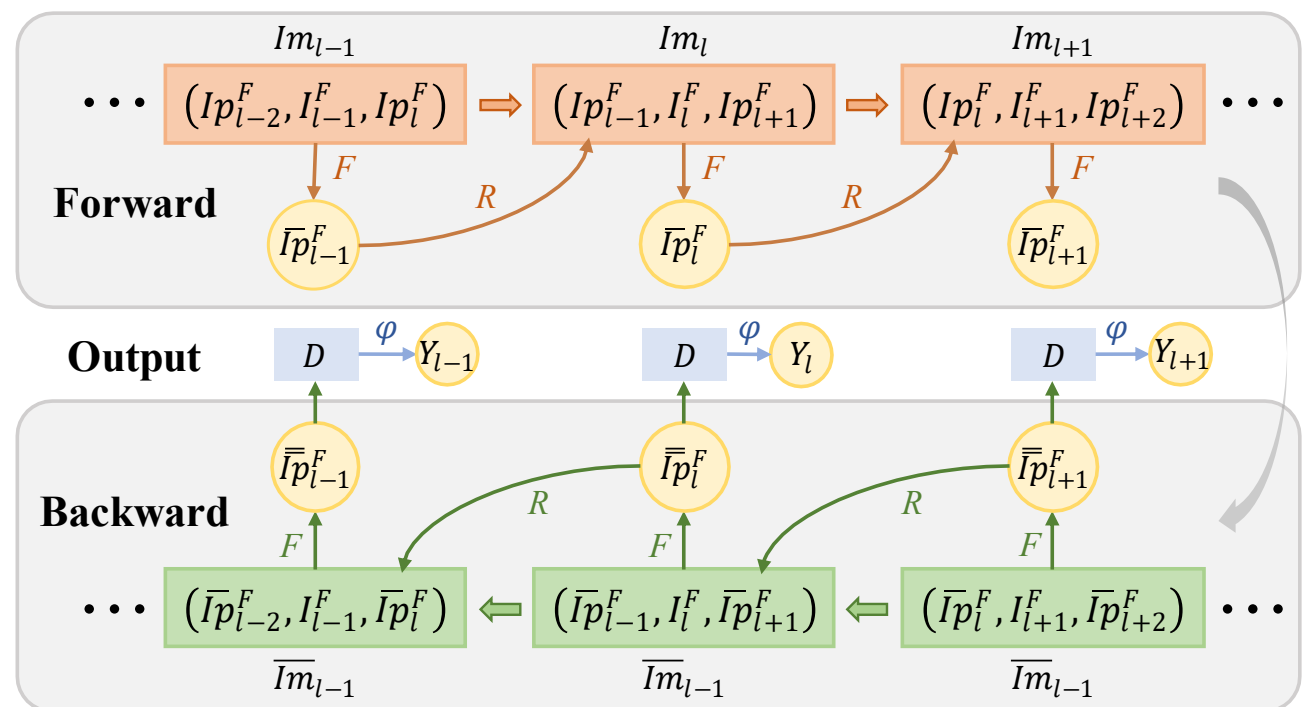

Fig. 4. The pipeline for the primary segmentation stage with a bi-directional recurrent update scheme (where 3 channels are used in this case).

**Forward Flow**: We present the update process of the forward flow starting with $Im_{l-1}$. First, the hybrid multi-channel data $Im_{l-1}$ is fed into the segmentation model $F$ to obtain the probabilistic map $\overline{Ip}^F_{l-1}$:

$$\overline{Ip}^F_{l-1} = F\left(Im_{l-1}\right) \tag{8}$$

Subsequently, $\overline{Ip}^F_{l-1}$ is applied to update the corresponding probabilistic map in the next set of data $Im_l$ to get $\overline{Im}_l$:

$$\overline{Im}_l = S\left(R\left(\overline{Ip}^F_{l-1}, Ip^F_{l-1}\right), I^F_l, Ip^F_{l+1}\right) \tag{9}$$

where $R$ denotes the updating process for the probabilistic map in $Im_l$, which can be formulated as:

$$R\left(\overline{Ip}^F_{l-1}, Ip^F_{l-1}\right) = \lambda_{new}\overline{Ip}^F_{l-1} + \lambda_{ori}Ip^F_{l-1} \tag{10}$$

where $\lambda_{new}$ and $\lambda_{ori}$ denote the weights of the new probabilistic map and the initial estimated probabilistic map, respectively, and the sum of $\lambda_{new}$ and $\lambda_{ori}$ is equal to 1. Both of these parameters are set to 0.5 in our experiments. Thus, the process of obtaining the probabilistic map $\overline{Ip}^F_l$ can be summarized as follows:

$$\overline{Ip}^F_l = F\left(S\left(R\left(\overline{Ip}^F_{l-1}, Ip^F_{l-1}\right), I^F_l, Ip^F_{l+1}\right)\right) \tag{11}$$

We employ $\overline{Ip}_l^F = T(Im_{l-1}; I_l^F, I_{l+1}^F)$ to simplify Eq (11), the probabilistic map $\overline{Ip}_{l+1}^F$ can be formulated as follows:

$$\begin{aligned} \overline{Ip}_{l+1}^F &= T\left(\overline{Ip}_l^F; I_{l+1}^F, Ip_{l+2}^F\right) \\ &= T\left(T\left(\overline{Ip}_{l-1}^F; I_l^F, Ip_{l+1}^F\right); I_{l+1}^F, Ip_{l+2}^F\right) \end{aligned} \tag{12}$$

The update process of the probabilistic maps would be propagated along with the forward flow as seen in Eq (12), thus providing more accurate guidance information for each group of hybrid 3-channel data. However, the forward propagation only updates the guidance information from the previous slice, ignoring the temporal information contained in the next slice. Therefore, we invert the forward flow to obtain the reverse propagation path.

**Backward Flow**: After one forward propagation, we invert the direction for backward propagation, so the process of obtaining the probabilistic map $\overline{\overline{Ip}}_l^F$ in backward flow can be expressed as:

$$\begin{aligned} \overline{\overline{Ip}}_l^F &= T\left(\overline{Ip}_{l-1}^F, I_l^F; \overline{\overline{Ip}}_{l+1}^F\right) \\ &= T\left(\overline{Ip}_{l-1}^F, I_l^F; T\left(\overline{Ip}_l^F; I_{l+1}^F, \overline{\overline{Ip}}_{l+2}^F\right)\right) \end{aligned} \tag{13}$$

Our bi-directional update flow can propagate local 3D hybrid information to each slice, thus avoiding the loss of contextual information between adjacent slices. This ensures that the final output is guided by the fully integrated local optimal 3D hybrid information. Under the guidance of effective contextual information within a certain range, the burden of obtaining an optimal refined pancreas segmentation result is highly reduced. Finally, the probabilistic maps $\overline{\overline{Ip}}^F$ are binarized (indicated by $\varphi$) and de-cropped (indicated by $D$) to get the segmentation result $Y$.

## 2.5. *Basic architecture and parameter setting*

We adopt the FCN (Long et al., 2015), a widely used segmentation baseline, as our initial estimation model. The motivation for applying FCN is mainly that it requires smaller GPU memory compared to other powerful backbones (*e.g.*, UNet), especially when the input size of the initial estimation model is 512×512. Considering that most of the background can be removed in primary segmentation applying the outputs of the initial estimation model, alleviating the high GPU memory requirement, we employ the UNet with skip connections as our primary segmentation model. The encoder-decoder and skip-connection techniques are beneficial in improving the final output resolution, and accurately locating and differentiating the pancreas from surrounding tissues (Drozdzal et al., 2016, González et al., 2018). Fig. 5 shows the overall structure of our UNet, which is composed of a pair of encoder and decoder modules, where each module consists of four blocks. Each block in encoder is composed of residual blocks, while each block in decoder two ordinary convolutional layers. A pooling layer is added at the end of each block in the encoder module, while a deconvolutional layer is added at the start of each block in the decoder module. The proposed method is implemented with the Pytorch library and trained on an NVIDIA GeForce GTX 2080Ti GPU. The stochastic gradient descent (SGD) is adopted to optimize parameters with learning rate of 1e-5 (M-FCN), 1e-4 (2.5D UNet) and batch size of 1, according to the empirical values that were widely used in pancreas segmentation tasks (Xie et al., 2019, Li et al., 2020, Chen et al., 2019, Fang et al., 2019, Zhou et al., 2017).

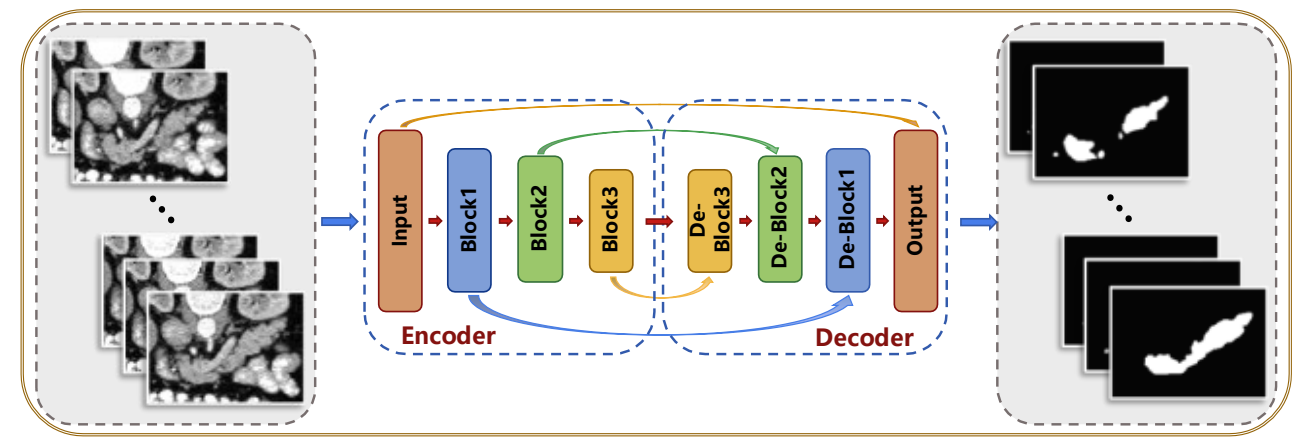

Fig. 5. The U-Net for bi-directional recurrent segmentation. The outputs are the pixel-wise probabilistic maps, which indicates the likelihood that the pixel belongs to the pancreas.

**Algorithm 1: Probabilistic Maps Guided Bi-directional Recurrent UNet**

**Input:** input volume $I$;
**Output:** segmentation volume $Y$;
1: $Ip \leftarrow E(I), I, Ip \in R^{m\times h\times w}$;
2: $I^F, Ip^F \leftarrow C(I, Ip; \varphi(Ip), m)$;
3: $Im \leftarrow S(I^F, Ip^F),\ Im \in R^{m\times h\times w\times c}$;
4: **bi-directional process:**
5: **for** $l \longleftarrow 1$ to $n$ **(forward):**
6: $\overline{Ip}_{l-1}^F \leftarrow F(Im_{l-1})$;
7: $\overline{Im}_l \leftarrow S\left(R\left(\overline{Ip}_{l-1}^F, Ip_{l-1}^F\right), I_l^F, Ip_{l+1}^F\right)$;
8: $\overline{Ip}_l^F \leftarrow F(\overline{Im}_l)$;
9: **end for**
10: **for** $l \longleftarrow n$ to 1 **(backward):**
11: $\overline{\overline{Ip}}_{l+1}^F \leftarrow F(\overline{Im}_{l+1})$;
12: $\overline{\overline{Im}}_l \leftarrow S\left(\overline{Ip}_{l-1}^F, I_l^F, R\left(\overline{\overline{Ip}}_{l+1}^F, Ip_{l-1}^F\right)\right)$;
13: $\overline{\overline{Ip}}_l^F \leftarrow F\left(\overline{\overline{Im}}_l\right)$;
14: **end for**
15: $Y \longleftarrow \varphi\left(D\left(\overline{\overline{Ip}}^F\right)\right)$;
16: **until** iteration termination condition reached;

## 2.6. *Loss function and inference procedure*

We employ dice similarity coefficient (DSC) loss for training and similarity characterization. Given the ground truth map $\hat{Y}$ and the final output $Y$, then the loss function can be defined as:

$$Loss\left(Y, \hat{Y}\right) = 1 - DSC\left(Y, \hat{Y}\right) = 1 - \frac{2\times\sum_i Y_i \hat{Y}_i}{\sum_i Y_i + \sum_i \hat{Y}_i} \tag{14}$$

The overall flow of our PBR-UNet is shown in Algorithm

1. Before the bi-directional recurrent segmentation, the M-FCN (denoted as $E$) is used to get the probabilistic maps $Ip$, which is then applied for cropping $I$ and $Ip$ to get $I^F$ and $Ip^F$. Next, image $I^F$ and the corresponding initial estimated probabilistic maps $Ip^F$ are combined by the transformation function $S$ to obtain the multi-channel data sample $Im$, which is then fed into the 2.5D UNet (denoted as $F$) for the bi-directional recurrent segmentation. This bi-directional recurrent segmentation is also performed along with three views, whose predicted probabilistic maps are then fused using arithmetic averaging. Finally, the predicted probabilistic maps are binarized and de-cropped to obtain the final output $Y$. For some cases with poor initial estimates, multiple forward and backward flows are required to obtain a stable result. Thus, we set an iteration termination condition, *i.e.*, the recurrent terminates when the similarity of two consecutive predictions exceeds a given threshold $\tau$ or when a fixed number of iterations $P$ is reached. Assuming that $Y_t$ and $Y_{t+1}$ are the predictions of the t-th and (t+1)-th iterations, respectively, we apply $DSC(Y_t, Y_{t+1})$ to calculate their similarity, and the iteration terminates when $DSC(Y_t, Y_{t+1}) > \tau$ or $t + 1 == P$. In our experiments, $\tau$ and $P$ are set to 0.95 and 5 respectively.

## 2.7. Evaluation metrics

We mainly use the DSC to evaluate the segmentation performance and apply average symmetric surface distance (ASD) and 95% Hausdorff distance (HD) to evaluate the results of modeling the inter-slice shape continuity. We also used the recall, precision, and intersection over union (IOU) to evaluate the pancreas localization performance. The IOU measure is the ratio of the intersection to the union of the predicted and real rectangular box, $IOU(Y, \hat{Y}) = |Box_Y \cap Box_{\hat{Y}}| \,/\, |Box_Y \cup Box_{\hat{Y}}|$. The recall indicates the proportion of the pancreas pixels that are correctly segmented to the total number of true pancreas pixels, $Recall(Y, \hat{Y}) = |Y \cap \hat{Y}| \,/\, \hat{Y}$. The precision reflects the proportion of the pancreas pixels that are correctly segmented to the total number of pixels that are labeled as a part of the pancreas, $Precision(X, \hat{Y}) = |Y \cap \hat{Y}| \,/\, Y$. Besides, we measured the standard deviation, maximum and minimum values, and then calculated the average of these metrics over all test cases.

# 3. Results

In this section, we conduct comprehensive experiments to analyse the effectiveness of our proposed PBR-UNet. We first introduce the performance of our proposed PBR-UNet (Sec. 3.1) and compare the results of different stages for ablation testing (Sec. 3.2). Then we introduce parameter selection (Sec. 3.3) and time consumption (Sec. 3.4), finally we compare our results with other state-of-the-art methods (Sec. 3.5).

## 3.1. The performance of our PBR-UNet

### 3.1.1 Segmentation performance

We quantitatively evaluated the segmentation results using the DSC, ASD, and HD indicators. The values of these indicators were reported in Table I, which showed that the mean DSC, ASD, and HD values of our model reached 85.35 ± 4.13%, 1.10 ± 0.40 mm, and 3.68 ± 2.30 mm, respectively. Fig. 6 visualized our segmentation results, most of the errors occurred at the pancreas edges, while the main part of the pancreas was correctly segmented. Fig. 7 showed the DSC distribution of 82 cases, where most of the results had a DSC value above 80%, which indicated that our method could improve the generalization performance to different CT volumes. This demonstrated a satisfactory performance of our method.

TABLE I
SEGMENTATION RESULT FOR THE NIH PANCREAS-CT DATASET.

| Evaluation | Min | Max | Mean | Std |
|---|---|---|---|---|
| DSC (%) | 71.36 | 91.05 | 85.35 | 4.13 |
| ASD (mm) | 0.66 | 2.95 | 1.10 | 0.40 |
| HD (mm) | 1.83 | 17.78 | 3.68 | 2.30 |

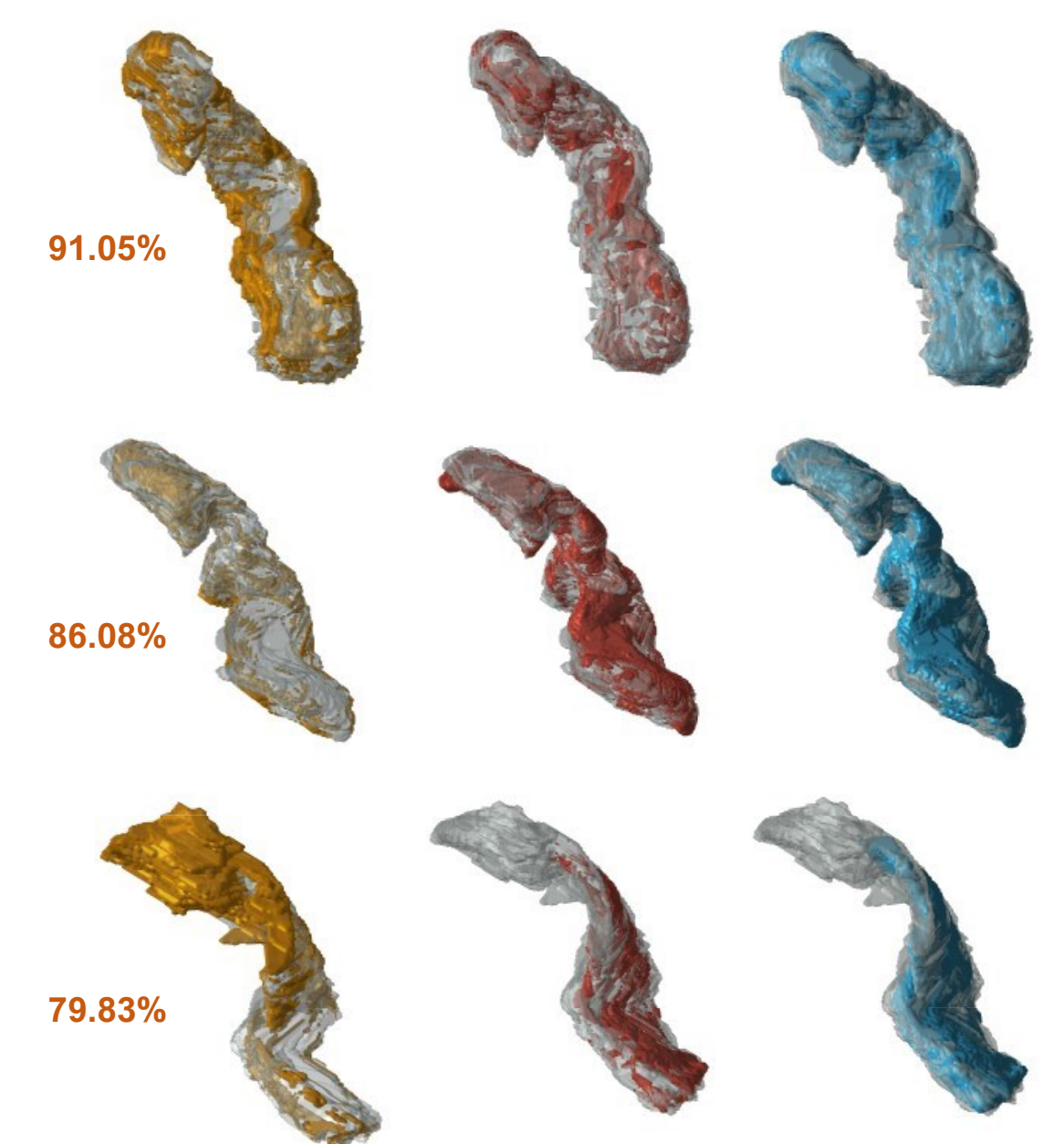

Fig. 6. Three-dimensional representations of the results compared to the labels (gray region). The yellow, red and blue region shows the under-segmentation, over-segmentation, and overall segmentation results, respectively.

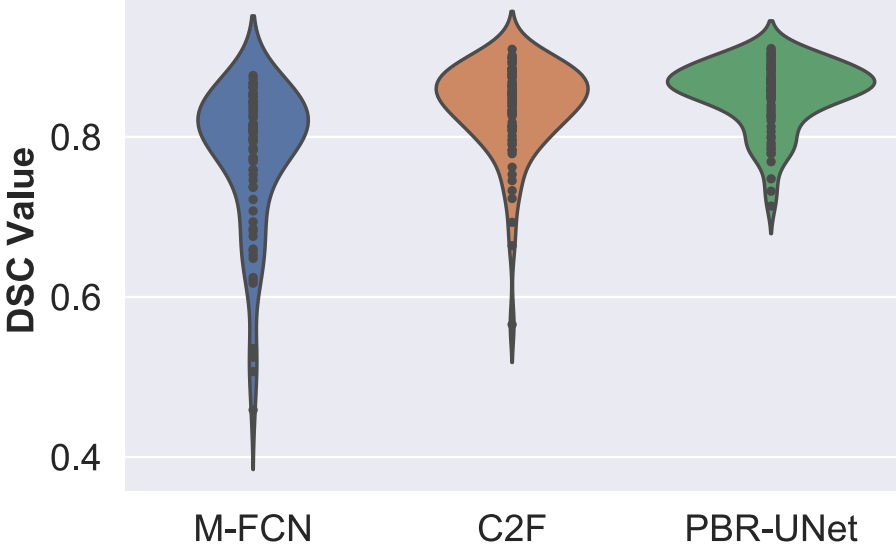

Fig. 7. The DSC distribution of 82 cases for the segmentation results.

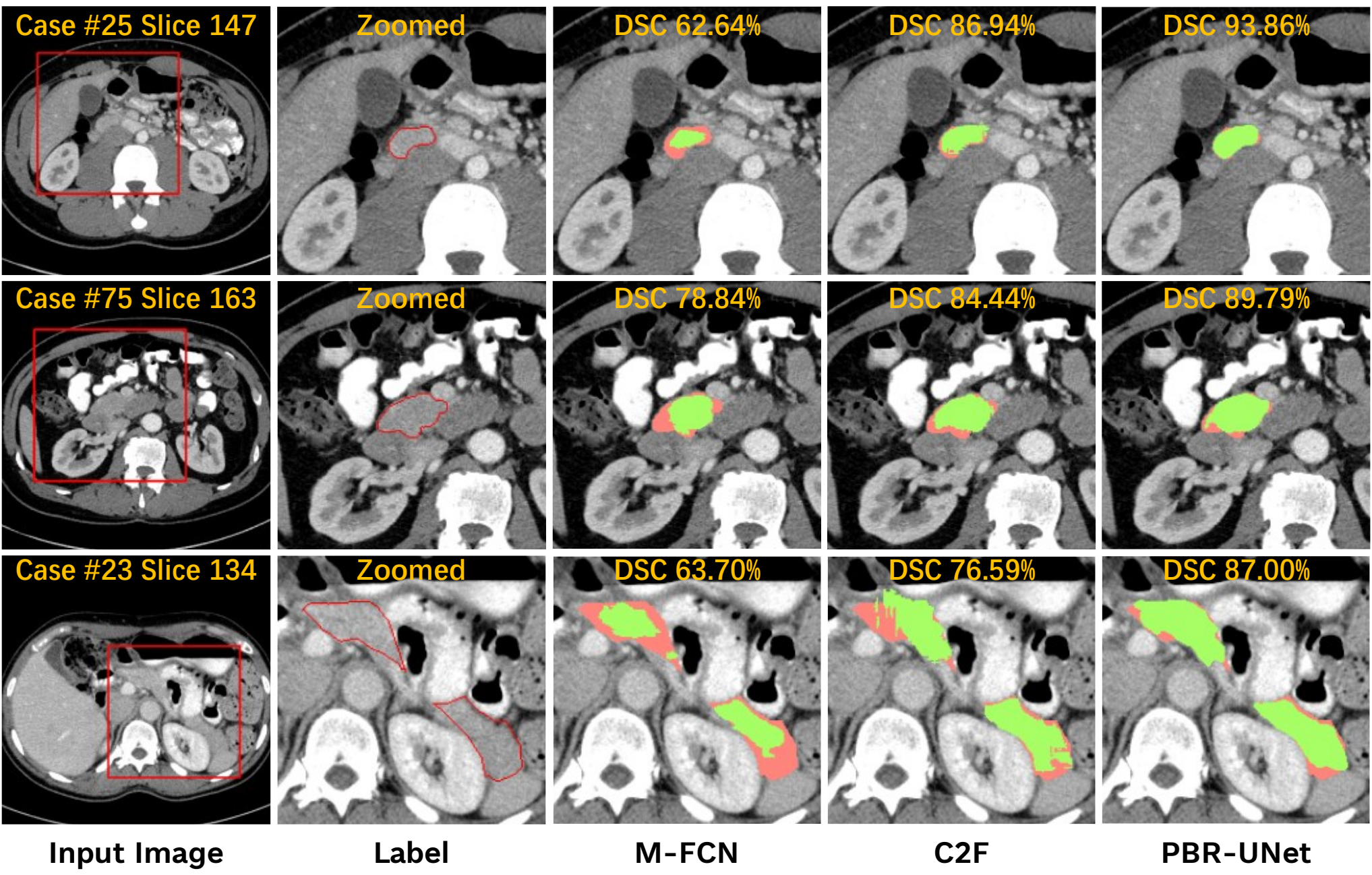

Fig. 8. Visual segmentation results for ablation experimental setting. The green region denotes the prediction, and the red region denotes the label.

### *3.1.2 Localization performance*

We applied the recall, precision, and IOU metrics to quantitatively evaluate the pancreas localization performance of our model. These three metrics were mainly focused on the shape and position of the segmented object. As shown in Table II, the recall, precision, and IOU reached 82.76%, 83.45%, and 75.03%, respectively. This showed that our model could accurately locate the pancreas in a global search.

TABLE II
THE PANCREAS LOCALIZATION PERFORMANCE.

| Evaluation | Min | Max | Mean | Std |
|---|---|---|---|---|
| Recall [%] | 54.06 | 94.41 | 82.76 | 8.21 |
| Precision [%] | 59.24 | 94.39 | 83.45 | 7.19 |
| IOU [%] | 53.33 | 86.58 | 75.03 | 7.46 |

## *3.2. Ablation test*

To justify the complementary roles of the initial estimation and the primary segmentation in our proposed framework, we compared the results of different stages (M-FCN, C2F, and PBR-UNet) in ablation experiments. Specifically, M-FCN denoted the initial estimation model, which consisted of three separate models segmenting along with different views. C2F was based on the general coarse-to-fine method. Fig. 8 demonstrated the visual improvement of PBR-UNet on the segmentation results. Moreover, most previous studies only focused on the overall performance of the model on each patient, neglecting the slice-wise results. Thus, the comparison of segmentation performance of different stages focused on two aspects, the volume-wise (Fig. 9 and Fig. 11) and the slice-wise (Fig. 10, Tables III, and Table IV).

### *3.2.1 Volume-wise segmentation performance*

Our evaluation of the volume-wise segmentation performance was reflected in three main aspects. First, we compared the overall performance of different models by showing the DSC distribution of all 82 cases (Fig. 7). Then, we approximated the reliability of the model by Fig. 9, which was defined as the proportion of results that are greater than a set criterion, with a larger proportion indicating that the model achieved the expected performance in a larger number of patients. Finally, to further assess the agreement between automated segmentation and human-guided segmentation results, we compared the volume correlation between expert annotations and segmentation results (Fig. 11). The slope of the linear regression curve represented the consistency of the model-based pancreas volume and the manually labeled reference volume. The closer the slope of the curve to 1 was, the better the segmentation result was.

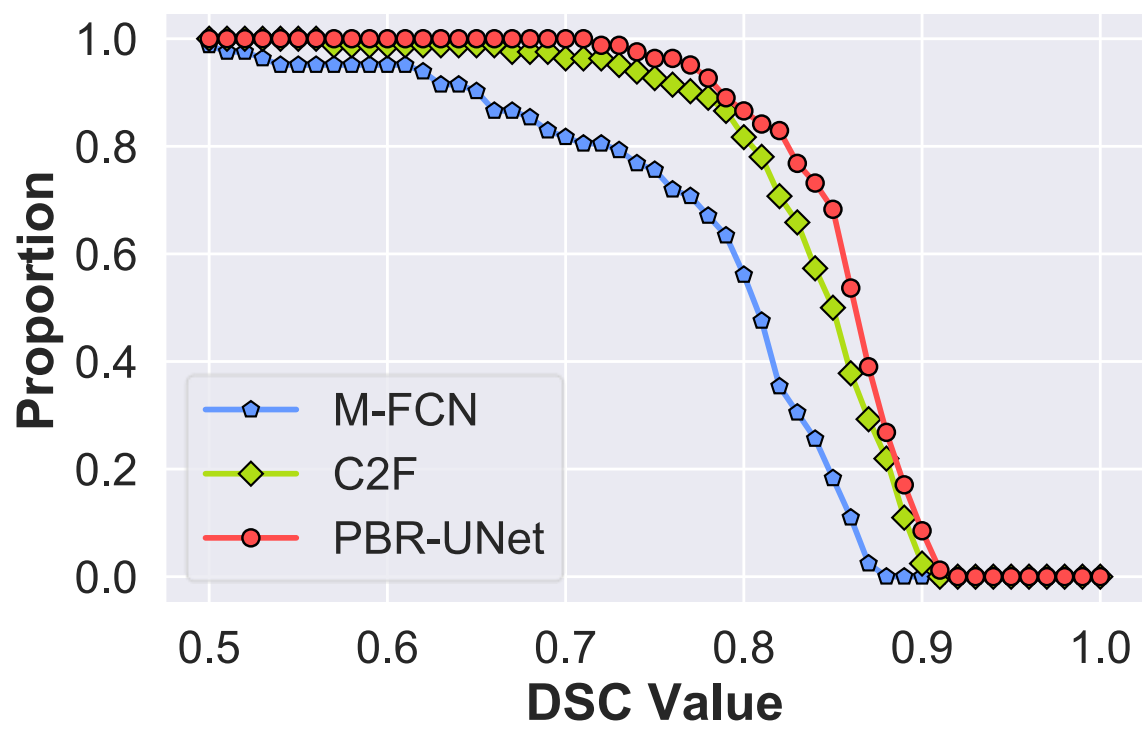

Fig. 9. The segmentation reliability in each ablation experimental setting for the analysis of the proposed model.

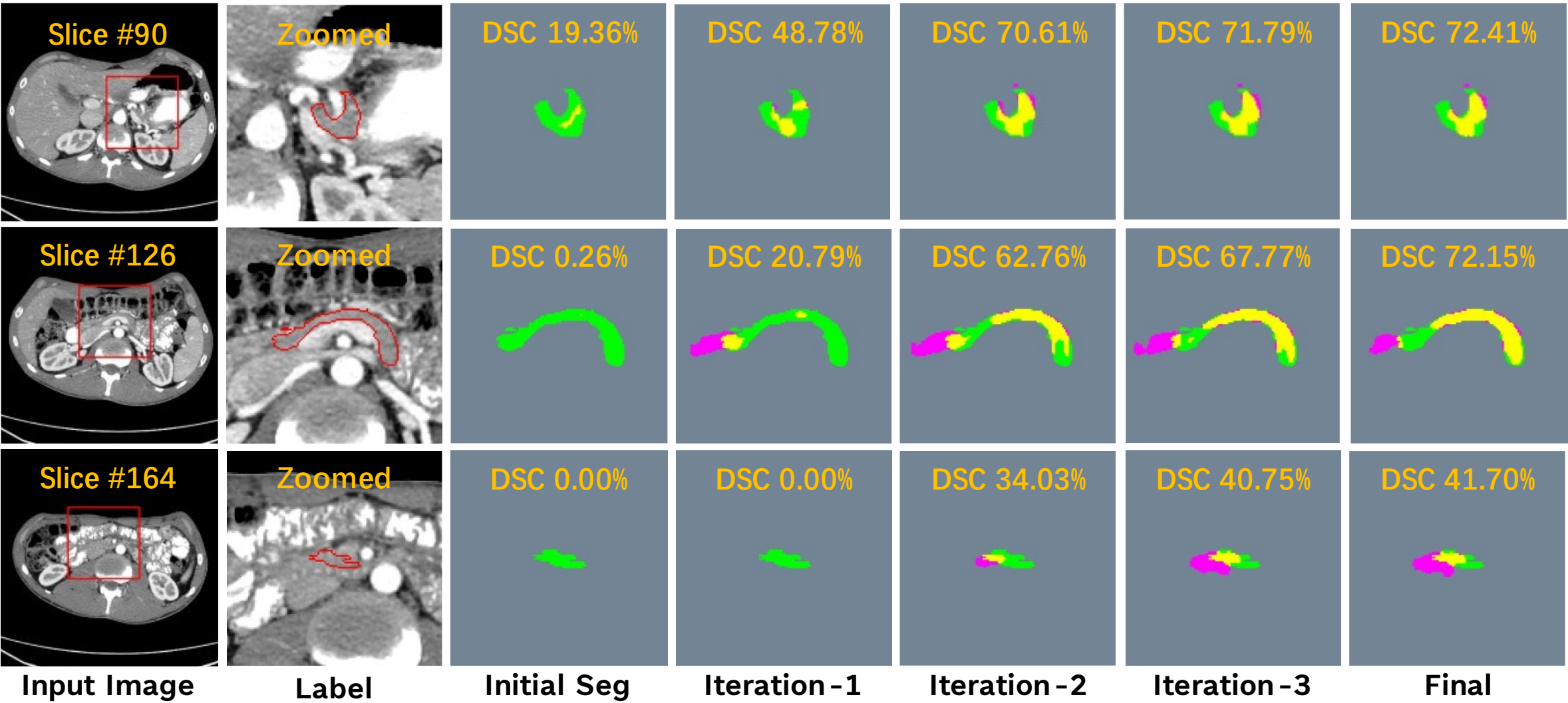

Fig. 10. Visual segmentation results of head, body, and tail of pancreas throughout the iteration process. The yellow, green, and purple regions denote the true positive, false negative, and false positive, respectively.

As shown in Fig. 7, the segmentation results of C2F were mostly concentrated at a DSC value of 0.82 compared to the dispersed results of M-FCN, which proved that the probabilistic maps provided by the initial estimation model successfully helped to achieve better performance and reflects the effectiveness of our initial estimation model. Compared to C2F, our proposed PBR-UNet achieved further performance improvement, resulting in a distribution of 82 results concentrated around 0.85, demonstrating the effectiveness of our proposed two-way cyclic update strategy. This gradual performance improvement was further reflected in Fig. 8. Moreover, Fig. 9 demonstrated the volume-wise reliability of our model, the curve of PBR-UNet was located at the top, indicating that it achieved the best results in different DSC intervals. As for the Bland-Altman agreement tests and regression curves in Fig. 11, PBR-UNet's results had the strongest correlation with manual labeling and most of the measurements are within ±1.96 standard deviations in Bland-Altman test. This further demonstrated the superiority of our model.

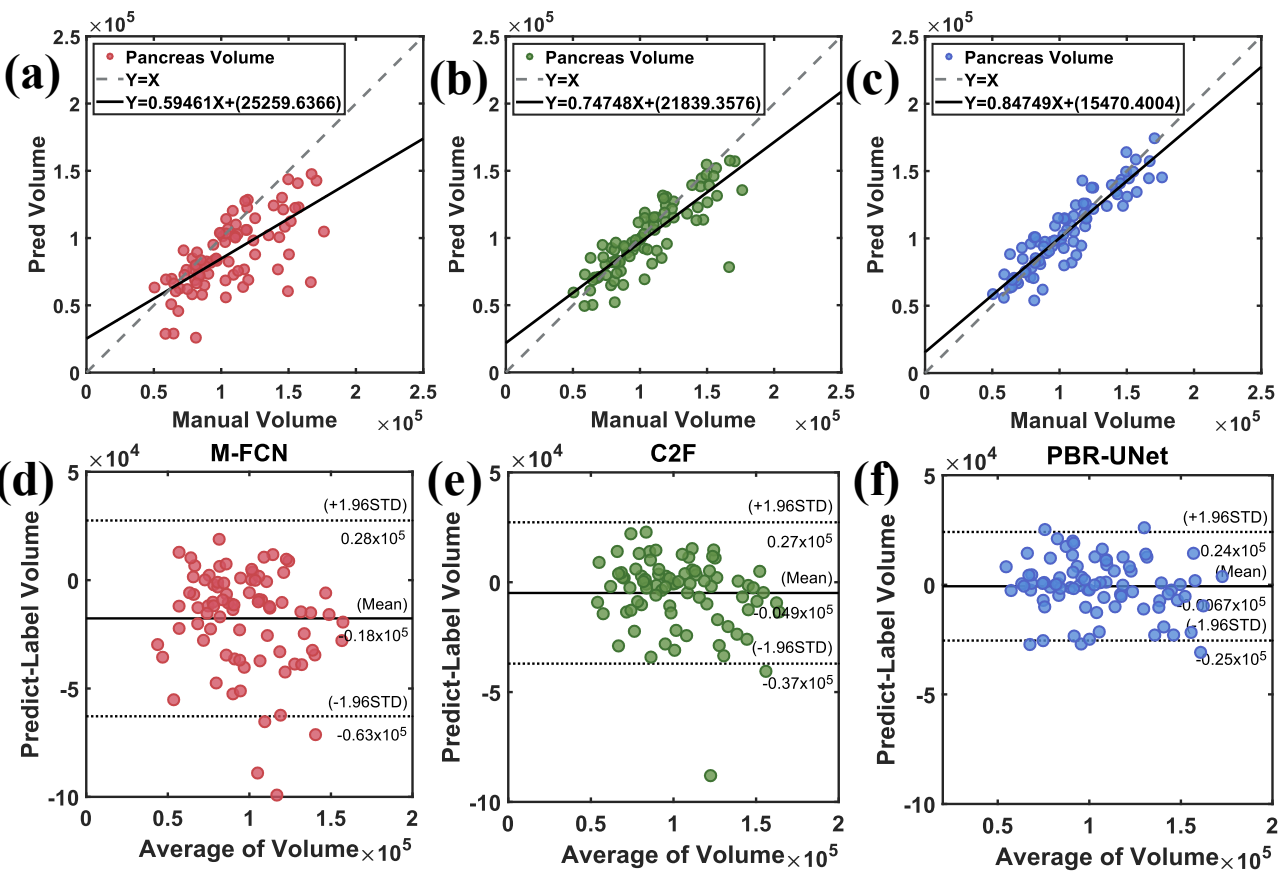

Fig. 11. Segmentation volumes of the M-FCN, C2F, and PBR-UNet versus the reference standard volumes. The first row shows the correlation of the segmentation volumes with the manually labeled volumes. The second row shows the results of the Bland-Altman agreement test which again compares between the manually segmented volumes and the segmentation volumes.

TABLE III
THE NUMBER OF SLICES IN THE DSC INTERVALS FOR DIFFERENT STAGES.

| Interval | M-FCN | C2F | PBR-UNet |
|---|---|---|---|
| [0,0.5) | 1326 (17.73%) | 712 (9.52%) | 567 (7.58%) |
| [0.5,0.6) | 381 (5.09%) | 224 (2.99%) | 163 (2.18%) |
| [0.6,0.7) | 762 (10.19%) | 474 (6.33%) | 376 (5.02%) |
| [0.7,0.8) | 1669 (22.32%) | 1203 (16.08%) | 1036 (13.85%) |
| [0.8,0.9) | 2982 (39.88%) | 3585 (47.94%) | 3647 (48.77%) |
| [0.9,1) | 357 (4.77%) | 1279 (17.10%) | 1688 (22.57%) |

### *3.2.2 Slice-wise segmentation performance*

The slice-based segmentation performance was also important for the integrity of pancreas segmentation. We analysed the distribution of the number of slices in the DSC intervals for different models, as shown in Table III. In our experiments, we found that most of the poorly segmented slices had a small target, so we then showed the segmentation results of slices with small-sized targets in Table IV.

From Table III, the segmentation results of 71.34% of the slices of PBR-UNet were distributed in [0.8,1.0), and only 7.58% of the slices had a DSC value below 0.5. Compared to the DSC distributions of M-FCN and C2F architectures, our method successfully improved the segmentation performance for most slices, achieving an acceptable slice-wise DSC distribution. However, due to the small size, we could still find that poorly segmented slices accounted for a small percentage of the results of our method. To observe the effect of pancreas size on segmentation performance, we calculated the segmentation results for slices of the head and tail of the pancreas and slices with pancreas size less than 300 pixels, respectively, in Table IV. The segmentation results suffered a dramatic drop due to the small size in the selected slices. However, despite the lack of sufficient context information, our model could still use local 3D hybrid information to optimize the segmentation of the poorly segmented slices and achieve the highest average DSC. Moreover, the number of slices with a DSC segmentation result of 0 was reduced by more than one-half compared to M-FCN. Fig. 10 showed the segmentation results for the head, body, and tail of the pancreas, respectively. As the iteration proceeds, the results of this three slices with failed initial segmentation were significantly improved. Overall, our approach could significantly improve the slice-wise performance in terms of segmentation small-sized targets.

TABLE IV
THE DSC OF THE SEGMENTATION RESULTS FOR SLICES WITH SMALL PANCREAS SIZES.

| Methods | Head and tail, 480 cases | | size ≤ 300, 496 cases | |
|---|---|---|---|---|
| | Failed slices | DSC | Failed slices | DSC |
| M-FCN | 128 | 36.23 | 165 | 33.25 |
| C2F | 75 | 50.53 | 98 | 45.65 |
| PBR-UNet | 53 | 59.93 | 68 | 55.53 |

TABLE V
COMPARISON OF THE PBR-UNET SEGMENTATION RESULTS WITH DIFFERENT GUIDANCE DEPTHS.

| Depth | DSC [%] | ASD [mm] | HD [mm] |
|---|---|---|---|
| 1 | 84.90 ± 4.27 | 1.11 ± 0.31 | 3.66 ± 1.79 |
| 2 | 85.35 ± 4.13 | 1.10 ± 0.40 | 3.68 ± 2.30 |
| 3 | 85.38 ± 4.27 | 1.10 ± 0.35 | 3.69 ± 2.57 |

### *3.3. Parameter selection*

The range of contextual information used for guiding the segmentation was limited to a fixed depth. We conducted comparative experiments to explore the impact of the guidance depth on the segmentation results. We set the navigation depth to 1, 2, and 3, respectively, while the channels of the multi-channel data would be 3, 5, and 7, respectively. All experiments were conducted under the same experimental settings. From Table V, PBR-UNet with a guidance depth of 3 achieved the best performance. Obviously, the results at a guidance depth of 2 and 3 were much better than those at a depth of 1. Increasing the guidance depth would enrich the local 3D hybrid information but would also introduce the interference from the sharply changing shape of the pancreas and increase resource consumption. Thus, to achieve a balance between accuracy and computational overhead, we employed a guidance depth of 2 for experiments.

### *3.4. The time consumption*

Our proposed bi-directional recurrent network based on probabilistic map guidance represented a lightweight solution, which had lower time resource consumption than the method of using 3D convolution to obtain context information. Since the time overhead depended not only on the algorithm but also on the hardware device, we compared the average running time of our algorithm at each stage based on the same device. Considering that different patients' volume data had a different number of slices, we calculated the inference time for each slice to quantify the model efficiency. As shown in Table VI, the time required for each stage was 0.17 s/slice, 0.38 s/slice, and 0.57 s/slice, respectively.

TABLE VI
TIME CONSUMPTION COMPARISON WITH DIFFERENT STAGES.

| Methods | Time[a] [s/slice] | | |
|---|---|---|---|
| | Min | Max | Mean |
| M-FCN | 0.12 | 0.19 | 0.17 ± 0.01 |
| C2F | 0.25 | 0.62 | 0.38 ± 0.06 |
| PBR-UNet | 0.36 | 1.21 | 0.57 ± 0.11 |

Time[a]: average time consumption for one axial slice image.

### *3.5. Comparison with state-of-the-art methods*

As shown in Table VII, our model exhibited a competitive performance against the state-of-the-art methods. Our method achieved a DSC value of 85.35%, which was close to the 85.53% obtained by (Khosravan et al., 2019). And (Fang et al., 2019) obtained the 85.46% DSC under the hold-out strategy (fixed 29 cases for testing), but the results based on the hold-out strategy were susceptible to the different test cases. Moreover, our method performed better than most 3D methods and 2D methods, demonstrating the effectiveness of our strategy of updating the local 3D hybrid information in a bi-directional recurrent.

Moreover, most pancreas segmentation approaches paid considerable attention to the overall segmentation performance, with little focus on the segmentation of small targets. However, due to the large shape and volume variability of the human pancreas, the ability to segment small-sized targets was an important indicator of segmentation stability. Therefore, we discussed the segmentation results of slices with small sizes in the above section on slice-wise analysis. Indeed, our model could optimize the segmentation results by propagating local 3D hybrid information. Figure 12 showed the segmentation results of five adjacent slices of the 6th patient. Due to the small size of the pancreas, the DSC value for the initial segmentation results was below 75%, and the closer to the tail, the smaller the pancreas size was. Since our method could propagate and update local contextual information in bi-directional recurrent segmentation, the results of all five slices achieved a DSC of more than 80% with the help of the hybrid information propagated from other slices. This emphasized that our model effectively used local 3D hybrid information, making large improvements in slices with poor segmentation.

TABLE VII
COMPARISON OF PANCREAS SEGMENTATION RESULTS WITH THE STATE-OF-THE-ART METHODS ON NIH DATA (DSC [%])

| Models | Method | Folds | DSC | Min | Max |
|---|---|---|---|---|---|
| (Asaturyan et al., 2019) | 3D | Hold-out | 79.30 ± 4.40 | 72.80 | 86.0 |
| (Roth et al., 2018a) | 2.5D | 4-CV | 81.27 ± 6.27 | 50.69 | 88.96 |
| (Oktay et al., 2018) | 3D | 5-CV | 81.48 ± 6.23 | / | / |
| (Zhou et al., 2017) | 2D | 4-CV | 82.37 ± 5.68 | 62.43 | 90.85 |
| (Li et al., 2020) | 2D | 4-CV | 83.06 ± 5.57 | 67.96 | 90.37 |
| (Guo et al., 2020) | 2D | Hold-out | 83.11 ± 3.06 | 75.38 | 88.61 |
| (Zheng et al., 2020) | 2D | 4-CV | 84.37 | / | / |
| (Xie et al., 2019) | 2.5D | 4-CV | 84.53 ± 5.30 | 64.40 | 90.93 |
| (Zhang et al., 2020) | 3D | 4-CV | 84.61 ± 5.21 | 70.36 | 91.46 |
| (Zhang et al., 2021) | 2.5D | 4-CV | 84.90 | 61.82 | 91.46 |
| (Chen et al., 2019) | 3D | 4-CV | 85.22 ± 4.07 | 71.40 | 91.36 |
| (Fang et al., 2019) | P3D | Hold-out | 85.46 ± 7.13 | 67.03 | 92.24 |
| (Khosravan et al., 2019) | 2D projections | 4-CV | 85.53 ± 1.23 | 83.20 | 88.71 |
| Ours | 2.5D | 4-CV | 85.35 ± 4.13 | 71.36 | 91.05 |

On the MSD dataset, our method also exhibited a competitive performance against the state-of-the-art methods.

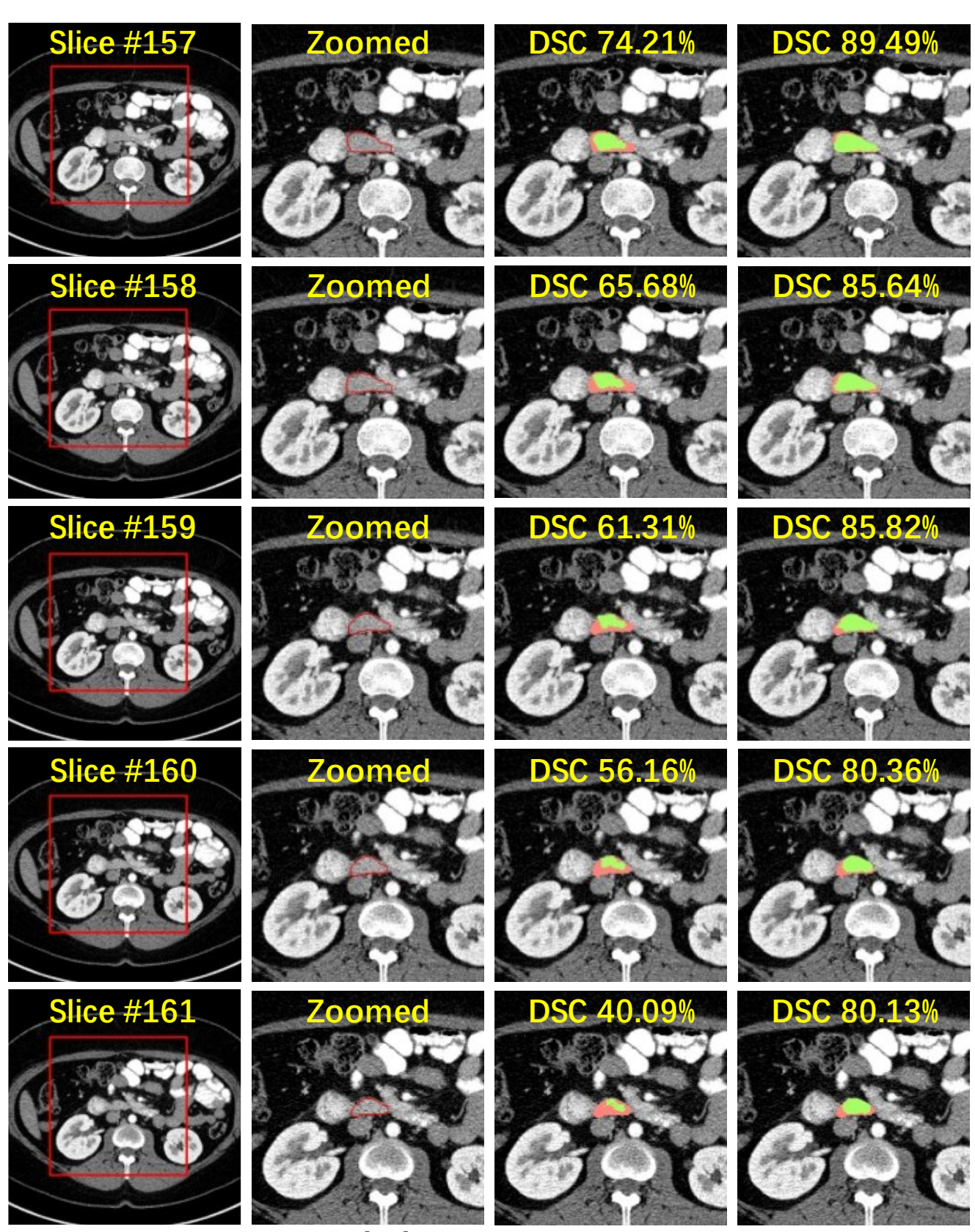

Fig. 12. The segmentation results of five adjacent slices with small target areas for 6th case. The green region denotes the prediction, and the red region denotes the label.

We applied the same strategy as (Zhang et al., 2021), used 211 cases for training and the remaining 70 cases for testing. To verify the effectiveness of our model on different data sets, all model parameters followed the settings on NIH. From Table VIII, we can observe that our method outperformed other state-of-the-art methods, obtaining 85.65% DSC of the results. Consequently, our method was demonstrated to achieve competitive performance on different datasets.

TABLE VIII
COMPARISON WITH STATE-OF-THE-ART METHODS ON MSD DATA (DSC[%]).

| Models | Mean | Min | Max |
|---|---|---|---|
| (Fang et al., 2019) | 84.71 | 58.62 | 95.54 |
| (Zhang et al., 2021) | 85.56 | 69.16 | 93.01 |
| Ours | 85.65 | 69.33 | 93.98 |

## 4. Discussion

### *4.1. Comparison on computing resources*

Pancreas segmentation was of great significance for clinical computer-aided diagnosis. The segmentation results could provide accurate location and contour information of the pancreas. This paper presented a segmentation network guided by probabilistic maps, which aimed to extract local 3D hybrid information with lower computational and time resource consumption. Here, we compared the complexity with the 3D method (Chen et al., 2019) in terms of model parameters, since their results were closest to ours (85.22% with 4-CV). Specifically, the parameters of the 2.5D UNet that used to update the local context information was 75 M. (Chen et al., 2019) fused 3D features and 2D features to effectively capture contextual information by the fusion module consisting of 3D UNet (90M), Vgg16 (138M) and ResNet18 (11M). The total parameters of these three models were 239M, which was much more than our method.

### *4.2. Multi-view fusion on three views*

In Fig. 11a, we noticed that the linear regression curve of M-FCN showed a large shift, with a correlation coefficient of only 0.59. This indicated that the volume of M-FCN segmentation results was significantly different from the expert annotation. And most of the sample points were located below the regression line, indicating that the volume of most predicted results was smaller than the manual labeling. The reason for this result lied mainly in the merging mechanism of the segmentation results of the three views. The merging method in this paper was a weighted average method. Hence, a pixel could be regarded within the pancreas area only when two views asserted their presence within the pancreas area. So, many uncertain pixels were discarded after averaging. Nevertheless, this mechanism provided higher confidence in the surviving pixels, thus avoiding the introduction of pixels with high uncertainty into the guidance information of the bi-directional scheme. Overall, three views fusion was a valuable and practicable initial estimation scheme for the 2D model, as verified by our experiments.

### *4.3. Probabilistic map guided bi-directional recurrent strategy*

Radiologists usually scroll up and down along one axis to locate the organ. Based on this practice, (Gonzalez et al., 2019) proposed a human behavior-driven method with prior knowledge. They introduced contours of other organs as prior information and used the current slice, the thresholded contour and the next slice to learn the connectivity between slices. By modeling segmentation behavior of humans, they achieved promising results. Nevertheless, our method still exhibited advantages in some aspects. Specifically, our method was performed in an end-to-end manner without manually introducing any prior knowledge. Besides, we applied probabilistic maps rather than thresholded contours to provide the distribution of the pancreas in adjacent slices, avoiding losing precious context. Moreover, we designed a bi-directional recurrent strategy to make full use of the context.

### *4.4. Limitation and future work*

Although our proposed method had achieved encouraging performance, there was still a shortcoming that need to be improved. Specifically, we found that a small number of ROIs obtained from the initial estimation did not cover all pancreatic regions, resulting in a limited performance improvement of these slices in the primary segmentation stage. This was the main problem faced by the stage-wise segmentation methods, *i.e.*, how to effectively use the initial results to provide a reasonable ROI. Thus, our future work will mainly focus on how to obtain a reasonable ROI in the stage-wise segmentation and avoid the adverse effects of a large or small ROI on the segmentation performance.

## 5. Conclusion

The current key challenge in pancreas segmentation is that 3D networks require high computing resources, while 2D networks cannot capture contextual information. Therefore, we propose a PBR-UNet for pancreatic segmentation. This network combines the intra-slice information and probabilistic maps of the adjacent slices into local 3D hybrid information for guiding segmentation. This information could be optimized by a bi-directional recurrent updating scheme. Extensive experiments have demonstrated that our proposed PBR-UNet method could achieve competitive results compared with other state-of-the-art methods. Besides, this new paradigm with the local 3D hybrid information and bi-directional recurrent updating scheme can be integrated with other segmentation models for compromising the trade-offs between 2D and 3D segmentation networks. Thus, our proposed PBR-UNet method not only provides a useful tool for pancreas segmentation but also a potential paradigm for research in medical image segmentation.

## Acknowledgements

This work was supported by the Innovation Research Plan from the Shanghai Municipal Education Commission (No. WF220408215), Med-Engineering Crossing Foundation from Shanghai Jiao Tong University (No. AH0820009), SJTU-Yitu Joint Laboratory of Artificial Intelligence in Healthcare funding (No. SA0820053/004) and Key Laboratory of Intelligent Computing in Medical Image, Ministry of Education.